\documentclass[runningheads,a4paper]{llncs}

\usepackage{amsmath}
\usepackage{amssymb}
\usepackage{graphicx}
\usepackage{cite}
\usepackage{url}
\usepackage{color} 
\urldef{\mailsa}\path|{olsonran}@upenn.edu}

\newcommand{\keywords}[1]{\par\addvspace\baselineskip
\noindent\keywordname\enspace\ignorespaces#1}

\newcommand{\be}{\begin{eqnarray}}
\newcommand{\ee}{\end{eqnarray}}
\begin{document}

\mainmatter  % start of an individual contribution

% first the title is needed
\title{Automating biomedical data science through tree-based pipeline optimization}

% a short form should be given in case it is too long for the running head
\titlerunning{Automating biomedical data science}

% the name(s) of the author(s) follow(s) next
%
% NB: Chinese authors should write their first names(s) in front of
% their surnames. This ensures that the names appear correctly in
% the running heads and the author index.
%
\author{Randal S.~Olson\inst{1}
\and Ryan J.~Urbanowicz\inst{1}
\and Peter C.~Andrews\inst{1}
\and Nicole A.~Lavender\inst{2}
\and La Creis Kidd\inst{2}
\and Jason H.~Moore\inst{1}}
\authorrunning{Randal S.~Olson et al.} % abbreviated author list (for running head)
% (feature abused for this document to repeat the title also on left hand pages)

% the affiliations are given next; don't give your e-mail address
% unless you accept that it will be published
\institute{Institute for Biomedical Informatics\\University of Pennsylvania\\3700 Hamilton Walk\\Philadelphia, PA 19104, USA\\
\email{olsonran@upenn.edu}
\and
University of Louisville\\505 S. Hancock St.\\Louisville, KY 40202, USA}

%\institute{Springer-Verlag, Computer Science Editorial,\\
%Tiergartenstr. 17, 69121 Heidelberg, Germany\\
%\mailsa}
%\mailsb\\
%\mailsc\\
%\url{http://www.springer.com/lncs}}

%
% NB: a more complex sample for affiliations and the mapping to the
% corresponding authors can be found in the file "llncs.dem"
% (search for the string "\mainmatter" where a contribution starts).
% "llncs.dem" accompanies the document class "llncs.cls".
%

%\toctitle{Lecture Notes in Computer Science}
%\tocauthor{Authors' Instructions}
\maketitle

\begin{abstract}

Over the past decade, data science and machine learning has grown from a mysterious art form to a staple tool across a variety of fields in academia, business, and government. In this paper, we introduce the concept of tree-based pipeline optimization for automating one of the most tedious parts of machine learning---pipeline design. We implement a Tree-based Pipeline Optimization Tool (TPOT) and demonstrate its effectiveness on a series of simulated and real-world genetic data sets. In particular, we show that TPOT can build machine learning pipelines that achieve competitive classification accuracy and discover novel pipeline operators---such as synthetic feature constructors---that significantly improve classification accuracy on these data sets. We also highlight the current challenges to pipeline optimization, such as the tendency to produce pipelines that overfit the data, and suggest future research paths to overcome these challenges. As such, this work represents an early step toward fully automating machine learning pipeline design.

\end{abstract}

\begin{figure}[t]
\begin{center}
\includegraphics[width=\textwidth]{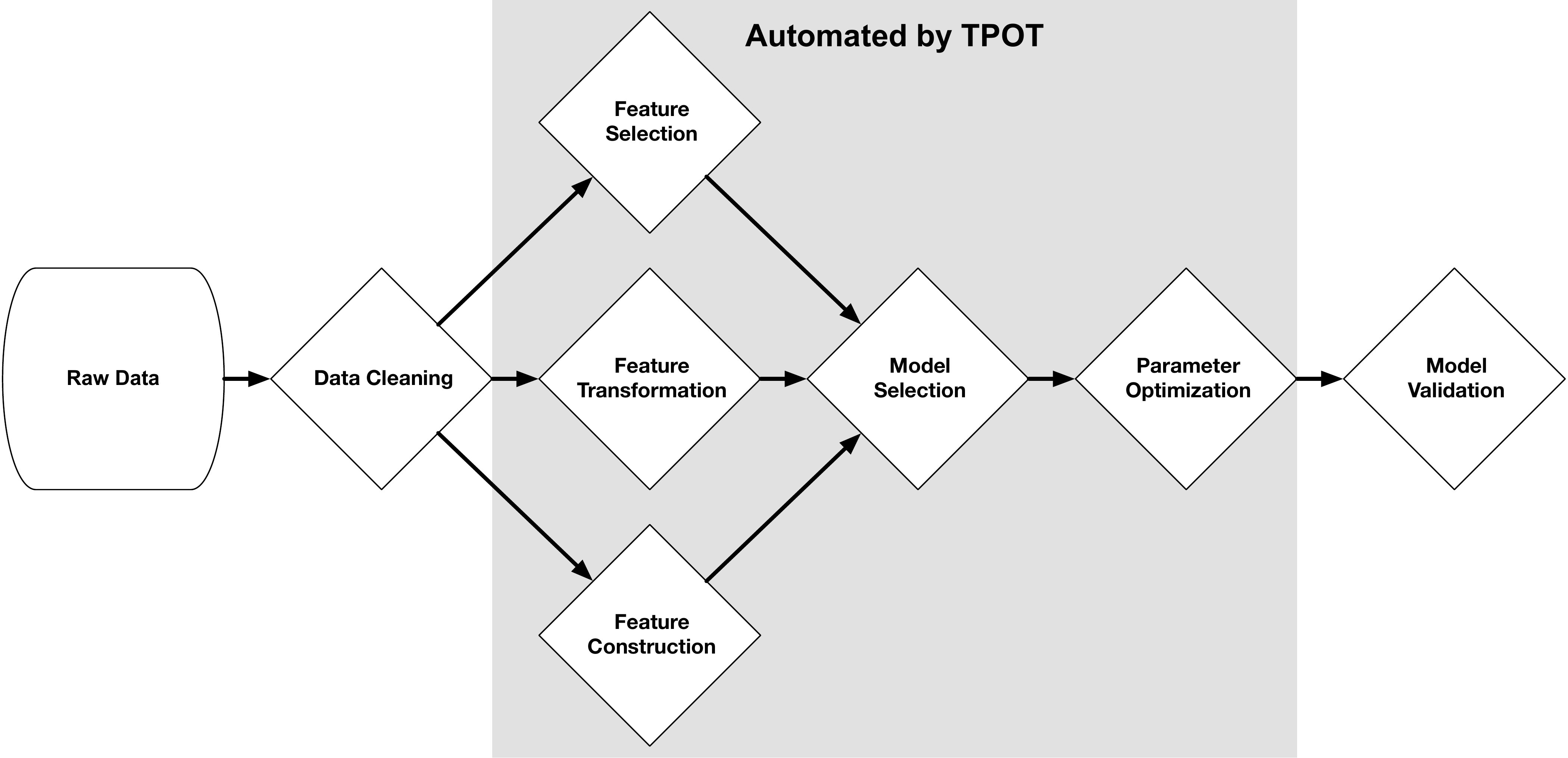}
\end{center}
\caption{
{\bf A depiction of a typical machine learning pipeline.} Before building a model of the data, the practitioner must ensure that the data is ready for modeling by performing an initial exploratory analysis (e.g., looking for missing or mislabeled data) and either correct or remove the offending records (i.e., data cleaning). Next, the practitioner may transform the data in some way to make it more suitable for modeling, e.g., by normalizing the features (i.e., feature transformation), removing features that do not seem useful for modeling (i.e., feature selection), and/or creating new features from the existing data (i.e., feature construction). Afterward, the practitioner must select a machine learning model to fit to the data (i.e., model selection) and select the model parameters that allow the model to make the most accurate classification from the data (i.e., parameter optimization). Lastly, the practitioner must validate the model in some way to ensure that the model's predictions generalize to data sets that it was not fitted on (i.e., model validation), for example, by testing the model's performance on a holdout data set that was excluded from the earlier phases of the pipeline. The light grey area indicates the steps in the pipeline that are automated by the Tree-based Pipeline Optimization Tool (TPOT).}
\label{fig:tpot-ml-pipeline-diagram}
\end{figure}

\section{Introduction}

Data science is a fast-growing field: Between 2011 and 2015, the number of self-reported data scientists has more than doubled~\cite{RJMetrics2015}. At the same time, machine learning---one of the primary tools of the modern data scientist---has experienced a revitalization as academics, businesses, and governments alike discovered new applications for automated algorithms that can learn from data. As a consequence, there has been a growing demand for tools that make machine learning more accessible, scalable, and flexible. Unfortunately, the successful application of these machine learning tools often requires expert knowledge of the tool as well as the target problem, an awareness of all assumptions involved in the analysis, and/or the application of simple exhaustive, brute force search. These requirements can make the application of many machine learning approaches a time-consuming and computationally-demanding endeavor.

As an example, a typical machine learning practitioner may build a pipeline as shown in Figure~\ref{fig:tpot-ml-pipeline-diagram}. At each step, there are dozens of possible choices to make: How to preprocess the data (e.g., what feature selector? what feature constructor? etc.), what model to use (e.g., support vector machine (SVM)? artificial neural network (ANN)? random forest (RF)? etc.), what parameters to use (e.g., how many decision trees in a RF? how many hidden layers in an ANN? etc.), and so on. Experienced machine learning practitioners often have good intuitions on what choices are appropriate for the problem domain, but some practitioners can easily spend several weeks tinkering with model parameters and data transformations until the pipeline achieves an acceptable level of performance.

In the past two decades, we have seen the growth of intelligent systems such as evolutionary algorithms that are capable of outperforming humans in a wide variety of tasks, such as antenna design for space missions~\cite{Hornby2011}, discovering and patching bugs in large software projects~\cite{Forrest2009}, and even in the study of finite algebras~\cite{Spector2008}. Considering the creative power of these intelligent systems, we must ask ourselves: Can intelligent systems automatically design machine learning pipelines?

In this paper, we report on the early development of an algorithm that automatically constructs and optimizes machine learning pipelines through a Tree-based Pipeline Optimization Tool (TPOT). We use a well-known evolutionary computation technique called genetic programming~\cite{Banzhaf1998} to automatically construct a series of data transformations and machine learning models that act on the data set with the goal of maximizing classification accuracy. We demonstrate TPOT's capabilities on an array of simulated data sets to explore the limits of the algorithm, then apply TPOT to a genetic analysis of prostate cancer. In particular, we show that TPOT can construct machine learning pipelines that achieve competitive classification accuracy, and that TPOT can discover novel pipeline operators---such as synthetic feature constructors---that significantly improve accuracy when added to the pipeline.

\section{Related Work}

Historically, automated machine learning pipeline optimization has focused on optimizing specific elements of the pipeline~\cite{Hutter2015}. For example, grid search is the most commonly-used form of hyperparameter optimization, where practitioners apply brute force to explore a broad ranged sweep of model parameter combinations in search of the parameter set that allows the machine learning model to perform best. Recent research has shown that randomly exploring parameter sets within the grid search often discovers high-performing parameter sets faster than an exhaustive search~\cite{Bergstra2012}, suggesting that there is promise for intelligent search in the hyperparameter space. Bayesian optimization of model hyperparameters, in particular, has been effective in this realm and has even outperformed manual hyperparameter tuning by expert practitioners~\cite{Snoek2012}. 

Another focus of automated machine learning has been feature construction. One recent example of automated feature construction is ``The Data Science Machine,'' which automatically constructs features from relational data sets via deep feature synthesis~\cite{Kanter2015}. In their work, Kanter {\em et al.} demonstrated the critical role of automated feature construction in machine learning pipelines by entering their Data Science Machine into three machine learning competitions and achieving expert-level performance in all of them.

All of these findings point to one take-away message: Intelligent systems are capable of automatically designing portions of machine learning pipelines, which can save practitioners considerable amounts of time by automating one of the most laborious parts of machine learning. To our knowledge, there have been no published attempts at automatically optimizing entire machine learning pipelines to date. Thus, the work presented in this paper establishes a blueprint for future research on the automation of machine learning pipeline design.

\section{Methods}

In this section, we describe tree-based pipeline optimization in detail, including the tools and concepts that underlie the Tree-based Pipeline Optimization Tool (TPOT). We begin this section by describing the basic pipeline operators that are currently implemented in TPOT. Next, we describe how the operators are combined together into a tree-based pipeline, and show how tree-based pipelines can be evolved via genetic programming. Finally, we end this section by providing an overview of the data sets that we used to evaluate TPOT.

\subsection{Decision Trees and Random Forests}

In the version of TPOT presented here, we only used decision tree and random forest machine learning models as they are implemented in scikit-learn~\cite{scikit-learn}, a general-purpose Python machine learning library. For both models, we used versions that perform binary classification, i.e., splitting the records into two pre-defined groups based on their feature vectors.

A {\em decision tree} model is a flowchart-like structure that asks a series of binary questions about each record's features, all the while attempting to differentiate the two groups as much as possible. An example question may be, ``Height $<=$ 182 cm?'', where the decision tree proceeds down the path to the right if the inequality is true and left otherwise. Generally, decision trees will select a question based on its ability to divide the records at each split by their group the most (i.e., maximizing the ``purity'' of the remaining records at that node), and will continue attempting to divide the remaining records down each path until either the maximum tree depth is reached or the path reaches a state where the remaining records completely belong to one group.

A {\em random forest} model uses several decision trees in an ensemble to make the same classification, where each decision tree is trained on a random sample (with replacement) of the training data. Once all of the decision trees in a random forest are constructed, their aggregate ``vote'' is used as the classification for the random forest. For further reading on decision trees and random forests, see~\cite{MachineLearningBook}.

\subsection{Synthetic Feature Construction}

In some cases, building new features from the existing feature set can prove useful for extracting vital information from data sets, especially when the features interact in some important way that would not be captured by methods that analyze only one feature at a time. In previous work, synthetic features constructed by random forests proved effective for combining genetic markers into constructed features that could then be used for classification~\cite{Pan2014}. In this paper, we allowed such synthetic features to be constructed by both decision trees and random forests. By default, whenever a decision tree or random forest was used to perform a classification in a TPOT pipeline, the classifications were also added as a constructed synthetic feature to the resulting data set.

\subsection{Decision Tree-based Feature Selection}

Often times, it is necessary to reduce the number of features in large data sets to improve classification accuracy---especially in genetic analyses, where it is not uncommon for the number of genetic features to number in the thousands or more. In this paper, we created a custom implementation of a decision tree-based feature selection method that reduces the feature set down to a parameterized number of feature pairs. To evaluate the feature pairs, we exhaustively constructed every possible two-feature combination from the feature set. These feature pairs were then ranked based on the training classification accuracy of a decision tree that was provided only those two features, where the feature pairs that resulted in higher training classification accuracy were selected first. This method allowed for the detection of epistatic interactions between features, which is typically overlooked by traditional machine learning methods that only consider the interaction between the endpoint and one feature at a time.

\subsection{Tree-based Pipelines}

\begin{figure}[t]
\begin{center}
\includegraphics[width=\textwidth]{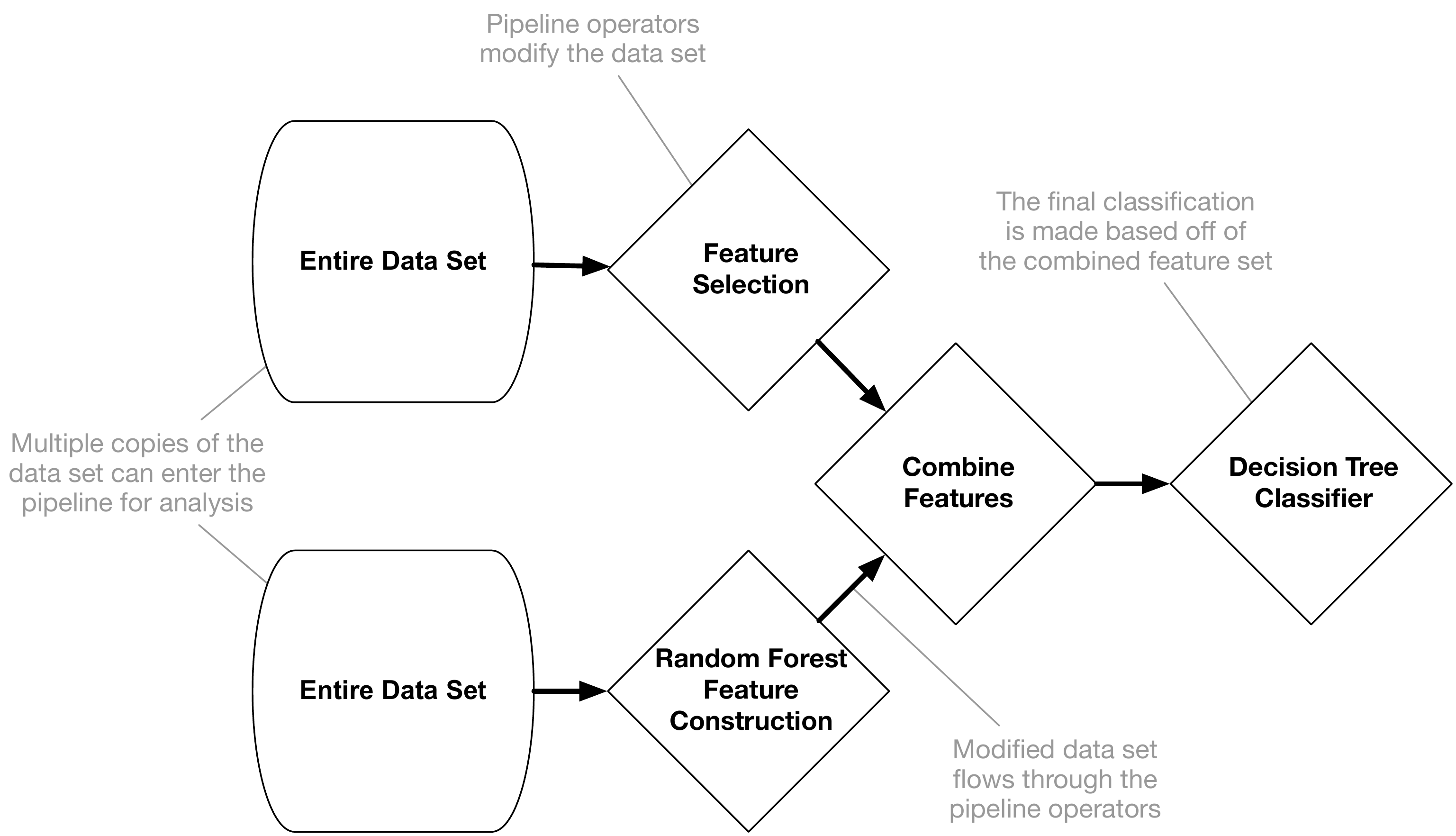}
\end{center}
\caption{
{\bf An example tree-based machine learning pipeline.} The data set flows through the pipeline operators, which add, remove, and modify the features in a successive manner. Combination operators allow separate copies of the data set to be combined, which can then be provided to a classifier to make the final classification.}
\label{fig:tpot-pipeline-example}
\end{figure}

To combine all of these operators into a flexible pipeline structure, we implemented the pipelines as trees as shown in Figure~\ref{fig:tpot-pipeline-example}. Every tree-based pipeline began with one or more copies of the input data set at the bottom of the tree, which was then fed into one of the many available pipeline operators: feature construction, feature selection, or classification by decision tree or random forest. These operators modified the provided data set then passed the resulting data set to the next operator as the data proceeded up the tree. In cases where multiple copies of the data set were being processed, it was also possible for the two data sets to be combined into a single data set via a data set combination operator.

Each time a data set passed through a decision tree or random forest operator, the resulting classifications were stored in a {\em guess} column, such that the most recent classifier to process the data would have its classifications in that column. Once the data set was fully processed by the pipeline---e.g., when the data set passed through the {\em Decision Tree Classifier} operator in Figure~\ref{fig:tpot-pipeline-example}---the values in the {\em guess} column were used to determine the classification accuracy of the pipeline. In this paper, we divided the data into stratified 75\% training and 25\% testing sets---such that the pipeline never trained but only predicted on the testing set---and each pipeline's accuracy was reported only on the testing set.

This tree-based pipeline structure allowed for arbitrary pipeline representations: For example, one pipeline could only apply operations in serial on a single copy of the data set, whereas another pipeline could just as easily work on several copies of the data set and combine them at the end before making a final classification.

\subsection{Genetic Programming}

To automatically generate these tree-based pipelines, we used a well-known evolutionary computation technique called genetic programming (GP) as implemented in the Python package DEAP~\cite{DEAP}. Traditionally, GP builds trees of mathematical functions to optimize toward a given criteria; in this case, by the same token, GP builds trees of pipeline operators to maximize the final classification accuracy of the pipeline. Here, we used GP to evolve the sequence of pipeline operators that acted on the data set as well as the parameters of these operators, e.g., the number of trees in a random forest or the number of feature pairs to select during feature selection.

In this paper, the TPOT GP algorithm followed a standard evolutionary algorithm procedure. At the beginning of every TPOT run, we randomly generated a fixed number of tree-based pipelines to constitute what we call the {\em population}. These pipelines were then evaluated based on their classification accuracy, which we used as the pipeline's {\em fitness}.

\begin{table}[t]
    \centering
    \caption{Genetic programming and experiment settings.}
    \begin{tabular}{l l}
        \hline \hline
        {\bf GP Parameter} & {\bf Value}\\ \hline
        Selection & 10\% elitism, rest 3-way tournament (2-way parsimony)\\
        Population size & 100\\
        Per-individual mutation rate & 90\%\\
        Per-individual crossover rate & 5\%\\
        Generations & 100\\
        Replicates & 30 \\
        \hline
    \end{tabular}
    \label{table:gp-settings}
\end{table}

Once all of the pipelines were evaluated, we proceeded to the next iteration (i.e., {\em generation}) of the GP algorithm. To generate the next generation's population, we first created exact copies of the pipeline with the highest fitness and placed them into the new population until they represented 10\% of the maximum population size (i.e., elitism of 10). To construct the remainder of the new population, we randomly selected three pipelines from the existing population then placed them in a tournament to decide which pipeline reproduces. In this tournament, the pipeline with the lowest fitness was removed, then the least complex pipeline of the remaining two (i.e., the pipeline with the fewest pipeline operators) was chosen as the winner to be copied into the new population (i.e., 3-way tournament selection with 2-way parsimony). This tournament selection procedure was repeated until the remaining 90\% of the new population was created.

With the new population created, we then applied a {\em one-point crossover} operator to a fixed percentage of the copied pipelines, where two pipelines are selected at random, split at a random point in the tree, then have their contents swapped between each other. Following that, a fixed percentage of the remaining unaffected pipelines had a random change (i.e., a {\em mutation}) applied to them:

\begin{itemize}
\item {\bf Uniform mutation}: A random operator in the pipeline was replaced with a new randomly-generated sequence of pipeline operators.
\item {\bf Insert mutation}: A new randomly-generated sequence of pipeline operators was inserted into a random place in the pipeline.
\item {\bf Shrink mutation}: A random subset of the pipeline operators were removed from the pipeline.
\end{itemize}
where each mutation operator had a $\frac{1}{3}$ chance of occurring when a pipeline was mutated. In all crossover and mutation operations, creating an invalid pipeline was disallowed, e.g., a pipeline that attempts to pass a data set into a parameter that expects a single integer would not be allowed.

Once the crossover and mutation operations completed, the previous generation's pipelines were deleted and this evaluate-select-crossover-mutate process was repeated for a fixed number of generations. In this manner, TPOT's GP algorithm continually tinkered with the pipelines---adding new pipeline operators that improve fitness and removing redundant or detrimental operators---in an intelligent, guided search for high-performing pipelines. At all times, the single best-performing pipeline ever discovered during the TPOT run was tracked and stored in a separate location and used as the representative pipeline at the completion of the run.

Table~\ref{table:gp-settings} describes the specific GP settings used in this paper. Every TPOT replicate was seeded with a unique random number generator seed to ensure that the runs were distinct.

\subsection{GAMETES Simulated Data Sets}

In order to evaluate TPOT, we adopted a diverse, complex simulation study design. We generated a total of 12 models and 360 associated data sets using GAMETES~\cite{Urbanowicz2012a}, an open source software package designed to generate a diverse spectrum of pure, strict epistatic models. GAMETES generated random, biallelic, n-locus single nucleotide polymorphism (SNP) models with a precise form of epistasis, which we refer to as pure. An {\em n}-locus model is purely epistatic if all {\em n} loci, but no fewer, are predictive of disease status. Models were precisely generated with the desired heritabilities, SNP minor allele frequencies, and population prevalences.

In this study, all data sets included 100 SNP attributes---8 that were predictive of a binary case/control endpoint, and 92 that were randomly generated using an allele frequency between 0.05 and 0.5. The 8 predictive SNPs were simulated as four separate purely epistatic models, additively combined using the newly-added ``hierarchical'' data simulation feature in GAMETES. In doing so, each separate interaction model additively contributed to the determination of the endpoint, but the overall data set did not include main effects (i.e., direct associations between single SNP variables and the endpoint).

We simulated two-locus epistatic genetic models with heritabilities of (0.1, 0.2, or 0.4) and attribute minor allele frequencies of 0.2 in GAMETES and selected the model with median difficulty from all those generated~\cite{Urbanowicz2012b}. Data sets with a sample size of either 200, 400, 800, or 1600 were generated, within which each of the four underlying two-locus epistatic models carried an equal additive weight. 30 replicates of each model and data set combination were generated, yielding a total of 360 data sets (i.e., 3 heritabilities * 4 sample sizes * 30 replicates). Together, this simulation study design allowed us to evaluate TPOT across a range of data sets with varying difficulties and sample sizes to explore the limits of TPOT's modeling capabilities.

\subsection{CGEMS Prostate Cancer Data Set}

To demonstrate TPOT's performance on a real-world data set, we applied TPOT to a genetic analysis of a nationally available genetic data set from 2,286 men of European descent (488 non-aggressive and 687 aggressive cases, 1,111 controls) collected through the Prostate, Lung, Colon, and Ovarian (PLCO) Cancer Screening Trial, a randomized, well-designed, multi-center investigation sponsored and coordinated by the National Cancer Institute (NCI) and their Cancer Genetic Markers of Susceptibility (CGEMS) program. We focused here on prostate cancer aggressiveness as the endpoint, where the prostate cancer is considered aggressive if it was assigned a Gleason score $\geq$ 7 and was in tumor stages III/IV. Between 1993 and 2001, the PLCO Trial recruited men ages 55-74 years to evaluate the effect of screening on disease specific mortality, relative to standard care. All participants signed informed consent documents approved by both the NCI and local institutional review boards. Access to clinical and background data collected through examinations and questionnaires was approved for use by the PLCO. Men were included in the current analysis if they had a baseline PSA measurement before October 1, 2003, completed a baseline questionnaire, returned at least one Annual Study Update (ASU), and had available SNP profile data through the CGEMS data portal (http://cgems.cancer.gov/). We used a biological filter to reduce the set of genes to just those involved in apoptosis (programmed cell death), DNA repair, and antioxidation/carcinogen metabolism. These biological processes are hypothesized to play an important role in prostate cancer. This report evaluated a total of 219 SNPs in the aforementioned biological pathways in relation to aggressive prostate cancer.

\begin{figure}
\begin{center}
\includegraphics[width=\textwidth]{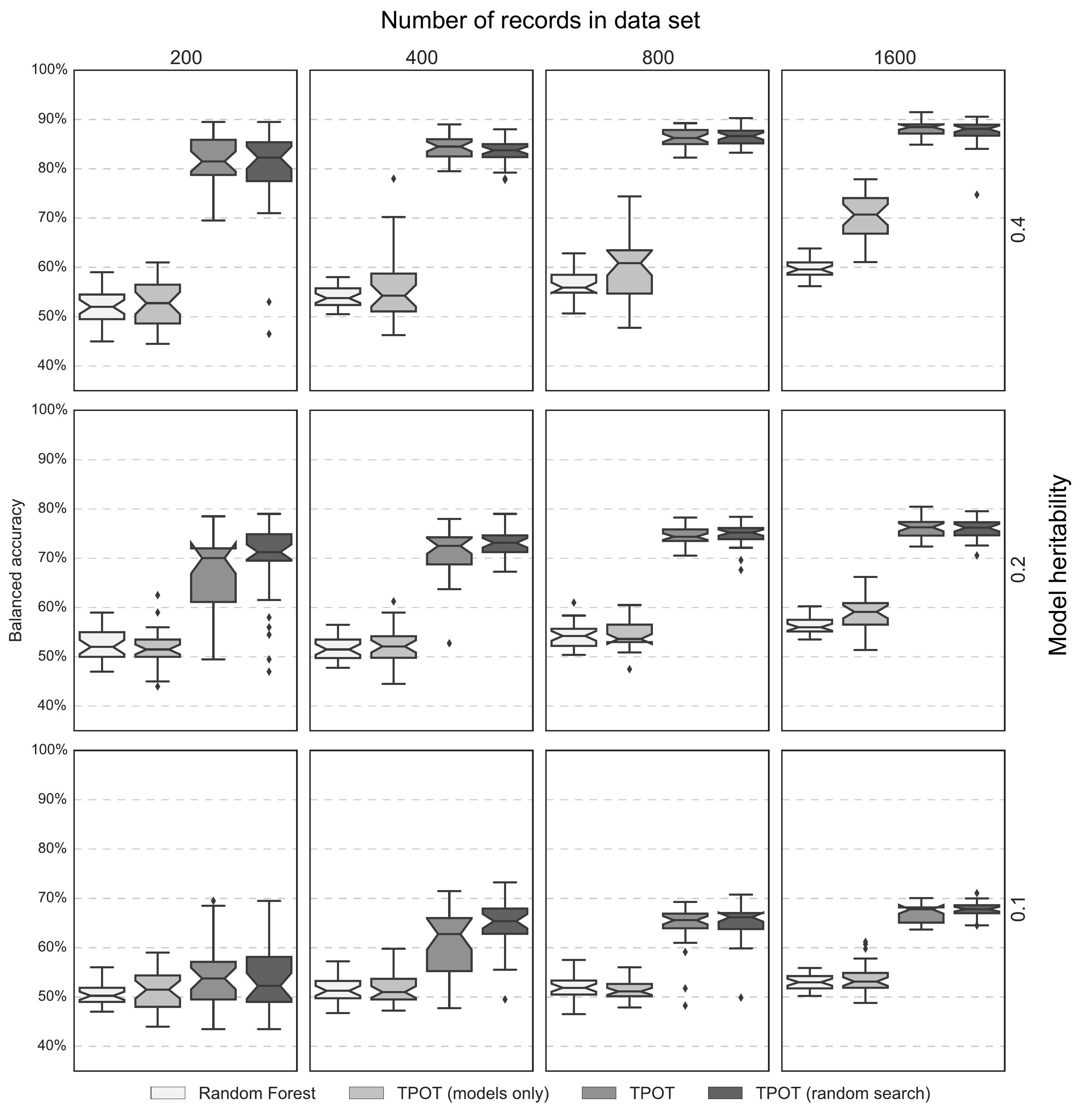}
\end{center}
\caption{
{\bf Tree-based Pipeline Optimization Tool (TPOT) performance comparison across a range of data set sizes and difficulties.} Each subplot on the grid shows the distribution of balanced 10-fold cross validation accuracies, where each notched box plot represents a sample of 30 data sets. (Note: The notches in the box plots indicate 95\% confidence intervals of the median.) The experiments compared include a random forest with 100 decision trees (``Random Forest''), a version of TPOT that only automates model selection and model parameters (``TPOT (models only)''), standard TPOT with genetic programming (``TPOT''), and standard TPOT with random generation of pipelines (``TPOT (random search)''). The subplots correspond to varying GAMETES configurations, where the x-axis modifies the number of records in the data set and the y-axis modifies the heritability in the model (where higher heritability reduces the amount of noise and vice versa). The grid ranges from easy configurations in the top right (larger data sets generated from higher heritability models) to difficult configurations in the bottom left (smaller data sets generated from low heritability models).}
\label{fig:tpot-gametes-comparison}
\end{figure}

\section{Results}

In this section, we present the performance of the tree-based pipeline optimization tool (TPOT) on the GAMETES simulated data sets and the real-world CGEMS prostate cancer data set.

\subsection{GAMETES Simulated Data Sets}

To begin, we tested TPOT on a broad range of heritability settings and data set sizes from GAMETES to explore TPOT's limits. Figure~\ref{fig:tpot-gametes-comparison} shows that in the higher heritability setting of 0.4---with the least amount of noise in the data set---TPOT achieved $>$80\% testing accuracy even with only 200 records to train on. Even at the more difficult heritability setting of 0.1---with a large amount of noise in the data set---TPOT achieved $>$65\% testing accuracy with only 800 records for training. However, at data set sizes of $<=$200 records and 0.1 heritability, the pipelines discovered by TPOT didn't perform much better than chance (i.e., 50\% testing accuracy due to the balanced classes). Generally, this array of tests demonstrated that TPOT performs best when provided with a) larger data sets to train on and/or b) models with higher heritability (i.e., less noise), which is to be expected since both configurations entail that the signal in the data set is easier to detect.

We also compared TPOT's classification performance to that of a basic random forest with 100 decision trees and a version of TPOT that optimized only model selection and parameters (i.e., neither versions had access to the feature selection nor feature construction operators). As shown in Figure~\ref{fig:tpot-gametes-comparison}, TPOT performed significantly better with the feature selection and construction operators included in all but the most difficult data sets. This finding showed that TPOT can discover useful feature construction and selection operators that that improve classification accuracy over solely optimizing the model parameters.

Lastly, we compared TPOT's classification performance with selection (i.e., genetic programming optimizing for classification accuracy) to a version of TPOT that randomly generated the same number of pipelines (i.e., randomly generates population size * generations number of pipelines, or 10,000 pipelines in this case). Figure~\ref{fig:tpot-gametes-comparison} shows that TPOT with selection did not perform significantly differently than TPOT via random search. This finding suggests that randomly generating pipelines eventually discovers a top-performing pipeline, and selection may not have performed a vital role on the GAMETES data sets.

\begin{figure}[t]
\begin{center}
\includegraphics[width=\textwidth]{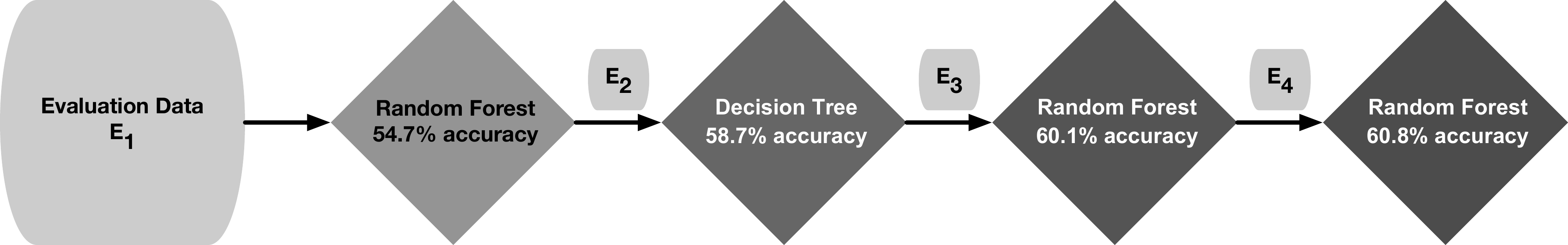}
\end{center}
\caption{
{\bf Balanced testing accuracy scores on the CGEMS prostate cancer data set as the data set progresses through the pipeline.} Each step in this diagram corresponds to a pipeline operator in the final TPOT pipeline. Starting from the raw data set, the pipeline applies a series of three feature construction steps (a random forest feature constructor, a decision tree feature constructor, then another random forest feature constructor), adding a new synthetic feature to the data set each time before proceeding to the next step. At the final classification step, the random forest classifier has access to a data set ($E_4$) containing all the original features along with the three new synthetic features, which allows it to accurately classify significantly more patients than without the synthetic features.}
\label{fig:tpot-cgems-accuracy-tree}
\end{figure}

\subsection{CGEMS Prostate Cancer Data Set}
\label{sec:results-cgems}

Next, we applied TPOT to the CGEMS prostate cancer data set to demonstrate its performance on a real-world genetic analysis. As shown in Figure~\ref{fig:tpot-cgems-accuracy-tree}, TPOT discovered a pipeline that achieved a balanced testing accuracy of 60.8\%, which is competitive with previous accuracy of 59.8\% on the same data set with the Computational Evolution System (CES)~\cite{Moore2013}. However, the balanced testing accuracy of the final TPOT pipeline dropped significantly to 51.7\% with 10-fold cross-validation, suggesting that there was overfitting occurring with the pipelines despite our use of a 75\%/25\% training/testing validation split during the optimization process.

In addition, Figure~\ref{fig:tpot-cgems-accuracy-tree} shows the progression of balanced testing accuracy as the pipeline added constructed features to the data set. Starting with the raw data set ($E_1$), a random forest is only capable of achieving a 54.7\% balanced testing accuracy on $E_1$. However, when the pipeline used that same random forest to construct and add a new synthetic feature to the data set---creating a new data set, $E_2$---a decision tree achieves a significantly-improved balanced testing accuracy of 58.7\%. At the final random forest classifier, the pipeline had added three synthetic features in total to the data set ($E_4$), enabling the random forest to achieve a much-improved 60.8\% balanced testing accuracy. These findings further support the theory that TPOT can discover novel feature construction methods that improve the pipeline's classification accuracy.

\begin{figure}
\begin{center}
\includegraphics[width=\textwidth]{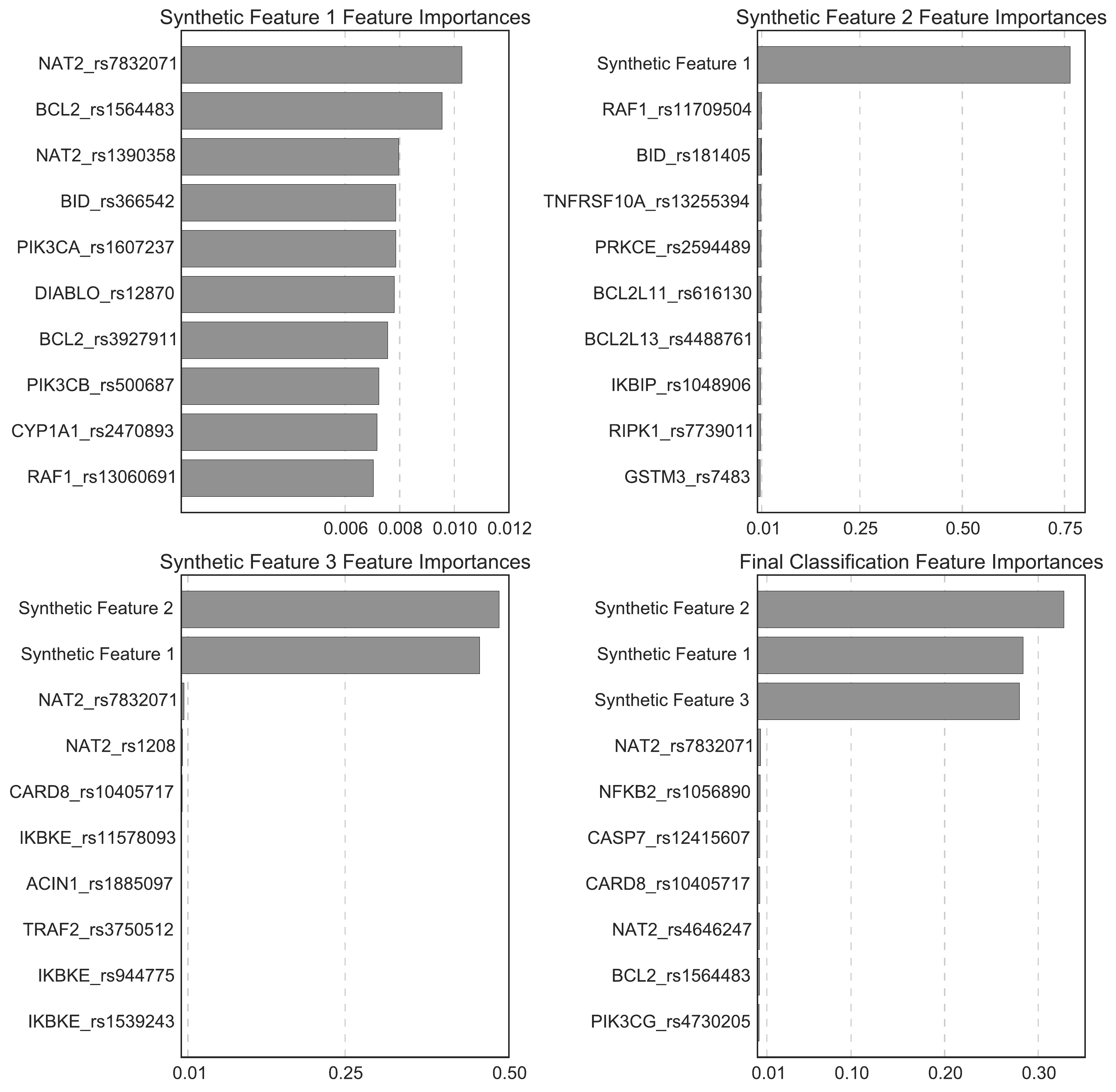}
\end{center}
\caption{
{\bf Top 10 features for each pipeline step for the CGEMS prostate cancer data set.} Each bar chart shows the feature importance scores of the top 10 features for each pipeline step, where the feature importance scores are determined by the feature's Gini importance~\cite{GiniImportance}, i.e., the sum of the feature's contributions to dividing the two classes. Synthetic Features 1, 2, and 3 are constructed features built by the pipeline operators. Each feature is named after the corresponding SNP in the CGEMS prostate cancer data set.}
\label{fig:tpot-cgems-feature-importances}
\end{figure}

To provide a more detailed view of the pipeline, we looked at the top 10 features (according to their feature importance scores) that were used during the synthetic feature construction and classification steps. When constructing the first synthetic feature (``Synthetic Feature 1'', Figure~\ref{fig:tpot-cgems-feature-importances}), the random forest primarily used the {\em NAT2} and {\em BCL2} SNPs to classify the patients. Interestingly, CES also discovered that these SNPs play an important role in determining the aggressiveness of prostate cancer~\cite{Moore2013}, which lends support to their importance in the biological functions of prostate cancer aggressiveness. During the construction of the second synthetic feature (``Synthetic Feature 2'', Figure~\ref{fig:tpot-cgems-feature-importances}), Synthetic Feature 1 played a much larger role in classifying the patients, further showing the synthesizing power of synthetic features. Despite the major role that Synthetic Feature 1 played, the decision tree was also able to integrate information from other SNPs such as {\em RAF1} and {\em BID} into Synthetic Feature 2. The third and final synthetic feature (``Synthetic Feature 3'', Figure~\ref{fig:tpot-cgems-feature-importances}) similarly relied on the previous synthetic features to classify the patients yet integrated new information from the remaining SNPs, which allowed the final random forest classifier to make a significantly improved classification based on the three synthetic features (``Final Classification'', Figure~\ref{fig:tpot-cgems-feature-importances}).

\section{Discussion}

It is important to note that the goal of pipeline optimization is not to replace data scientists nor machine learning practitioners. Rather, we aim for the tree-based pipeline optimization tool (TPOT) to be a ``Data Science Assistant'' that can explore the data, discover novel features, and recommend pipelines to the practitioner. From there, the practitioner is free to build on the automated pipelines and integrate their domain knowledge as they see fit. To aid in this goal, we have released TPOT as an open source Python package that provides a flexible implementation of the concepts introduced in this paper. We encourage interested practitioners to involve themselves in the project on GitHub (https://github.com/rhiever/tpot/).

In this paper, we showed that TPOT is capable of building machine learning pipelines that achieve competitive classification accuracy and discovering novel pipeline operators---such as synthetic feature constructors---that significantly improve classification accuracy. However, TPOT is not without its drawbacks. For example, TPOT with guided search did not perform significantly differently than TPOT with randomly-generated pipelines (Figure~\ref{fig:tpot-gametes-comparison}), which suggests that the additional layer of guided search (i.e., genetic programming) is not useful in the current version of TPOT. We believe that guided search did not outperform random search in this case because TPOT currently lacks clear building blocks~\cite{Goldberg2002}---i.e., small combinations of pipeline operators that can be combined to improve accuracy---for evolution to act on, and instead builds pipelines from powerful pipeline operators such as random forest classifiers. In future work, we will explore methods to divide the pipeline operators into better building blocks for TPOT to work with.

Furthermore, in Section~\ref{sec:results-cgems} we showed that TPOT built a pipeline that suffered significantly from overfitting on the CGEMS prostate cancer data set. In this work, we attempted to prevent overfitting by evaluating the TPOT pipelines by dividing the data set into a 75\%/25\% training/testing validation set, where the pipelines were trained on the training set and assigned a fitness according to their balanced accuracy on the testing set. In place of overfitting on the training data, the pipeline overfit on the testing data instead, which suggests that alternative methods are required to prevent TPOT from overfitting. In future work, we will explore methods such as multi-objective~\cite{Konak2006} and Pareto optimization~\cite{Deb2002}, where one of the objectives/Pareto fronts can be pipeline generalization performance or pipeline complexity (assuming less-complex pipelines will generalize better).

In the GAMETES simulated data sets, the random forest and decision tree classifiers failed to detect the epistatic pairs in the data sets on their own, which resulted in poor classification accuracy (Figure~\ref{fig:tpot-gametes-comparison}, ``Random Forest'' and ``TPOT (models only)''). This failure likely occurred because traditional machine learning algorithms only look at single feature-class correlations, which will necessarily overlook any interactions between the features. We believe that this result points to the need for machine learning algorithms such as Spatially Uniform ReliefF (SURF)~\cite{Greene2009} that can reliably and efficiently discover important features that have epistatic interactions with other features.

When applied to the CGEMS prostate cancer data set, TPOT discovered several synthetic features that significantly contributed to the classification accuracy of the pipeline. Notably, several of the SNPs that TPOT used to classify the patients---{\em NAT2} and {\em BCL2}, in particular---were also found to play an important role in the prediction prostate cancer aggressiveness in a previous study~\cite{Moore2013}, which highlights TPOT's ability to contribute to knowledge discovery. In future work, we will continue to develop methods to extract new knowledge such as feature importances from TPOT pipelines.

Beyond the work discussed above, in the near future we will continue developing TPOT to integrate more feature selection (e.g., SURF~\cite{Greene2009}) and construction operators, more machine learning models (e.g., support vector machines, logistic regression, K-nearest neighbor classifiers, etc.), and better hyperparameter optimization (e.g., Bayesian optimization methods~\cite{Snoek2012}). In general, we seek to integrate as many pipeline operators as possible into TPOT, with the assumption that genetic programming will discover the best operators (or set of operators) for the classification task at hand.

\section{Conclusions}

Machine learning pipeline optimization is poised to transform data science by automating one of the most tedious parts of machine learning. In this paper, we introduced a new method for automatically creating and optimizing machine learning pipelines, tree-based pipeline optimization, and demonstrated its effectiveness on a series of simulated and real-world genetic data sets. In particular, we showed that such a system can automatically build machine learning pipelines that achieve competitive classification accuracy as well as discover novel pipeline operators---such as synthetic feature constructors---that significantly improve classification accuracy on these data sets. We also highlighted the current challenges to pipeline optimization, such as the tendency to produce pipelines that overfit the data, and suggested future research paths to overcome these challenges. As such, this work represents an early step toward fully automating machine learning pipeline design.

% Do NOT remove this, even if you are not including acknowledgments
\section{Acknowledgments}

We thank Sebastian Raschka for his valuable input during the development of this project. We also thank the Michigan State University High Performance Computing Center for the use of their computing resources. This work was supported by National Institutes of Health grants LM009012, LM010098, and EY022300.

%\section*{References}
% The bibtex filename
\bibliography{references}

\begin{thebibliography}{10}

\bibitem{RJMetrics2015}
RJMetrics:
\newblock {The State of Data Science} (November 2015)
  \url{https://rjmetrics.com/resources/reports/the-state-of-data-science/}.

\bibitem{Hornby2011}
Hornby, G.S., Lohn, J.D., Linden, D.S.:
\newblock {Computer-automated evolution of an X-band antenna for NASA's Space
  Technology 5 mission.}
\newblock Evolutionary Computation \textbf{19}(1) (2011)  1--23

\bibitem{Forrest2009}
Forrest, S., Nguyen, T., Weimer, W., Le~Goues, C.:
\newblock A genetic programming approach to automated software repair.
\newblock In: Proceedings of the 11th Annual Conference on Genetic and
  Evolutionary Computation. GECCO '09, New York, NY, USA, ACM (2009)  947--954

\bibitem{Spector2008}
Spector, L., Clark, D.M., Lindsay, I., Barr, B., Klein, J.:
\newblock Genetic programming for finite algebras.
\newblock In: Proceedings of the 10th Annual Conference on Genetic and
  Evolutionary Computation. GECCO '08, New York, NY, USA, ACM (2008)
  1291--1298

\bibitem{Banzhaf1998}
Banzhaf, W., Nordin, P., Keller, R.E., Francone, F.D.:
\newblock {Genetic Programming: An Introduction}.
\newblock Morgan Kaufmann, San Meateo, CA (1998)

\bibitem{Hutter2015}
Hutter, F., L{\"u}cke, J., Schmidt-Thieme, L.:
\newblock {Beyond Manual Tuning of Hyperparameters}.
\newblock KI - K{\"u}nstliche Intelligenz \textbf{29}(4) (2015)  329--337

\bibitem{Bergstra2012}
Bergstra, J., Bengio, Y.:
\newblock {Random Search for Hyper-Parameter Optimization}.
\newblock Journal of Machine Learning Research \textbf{13} (2012)  281--305

\bibitem{Snoek2012}
Snoek, J., Larochelle, H., Adams, R.P.:
\newblock {Practical Bayesian Optimization of Machine Learning Algorithms}.
\newblock In Pereira, F., Burges, C.J.C., Bottou, L., Weinberger, K.Q., eds.:
  Advances in Neural Information Processing Systems 25.
\newblock Curran Associates, Inc. (2012)  2951--2959

\bibitem{Kanter2015}
Kanter, J.M., Veeramachaneni, K.:
\newblock {Deep Feature Synthesis:Towards Automating Data Science Endeavors}.
\newblock In: Proceedings of the International Conference on Data Science and
  Advance Analytics, IEEE (2015)

\bibitem{scikit-learn}
Pedregosa, F., Varoquaux, G., Gramfort, A., Michel, V., Thirion, B., Grisel,
  O., Blondel, M., Prettenhofer, P., Weiss, R., Dubourg, V., Vanderplas, J.,
  Passos, A., Cournapeau, D., Brucher, M., Perrot, M., Duchesnay, E.:
\newblock Scikit-learn: Machine learning in {P}ython.
\newblock Journal of Machine Learning Research \textbf{12} (2011)  2825--2830

\bibitem{MachineLearningBook}
Hastie, T.J., Tibshirani, R.J., Friedman, J.H.:
\newblock The Elements of Statistical Learning: Data Mining, Inference, and
  Prediction.
\newblock Springer, New York, NY, USA (2009)

\bibitem{Pan2014}
Pan, Q., Hu, T., Malley, J.D., Andrew, A.S., Karagas, M.R., Moore, J.H.:
\newblock {A System-Level Pathway-Phenotype Association Analysis Using
  Synthetic Feature Random Forest}.
\newblock Genetic Epidemiology \textbf{38}(3) (2014)  209--219

\bibitem{DEAP}
Fortin, F.A., {De Rainville}, F.M., Gardner, M.A., Parizeau, M., Gagn\'e, C.:
\newblock {DEAP: Evolutionary Algorithms Made Easy}.
\newblock Journal of Machine Learning Research \textbf{13} (2012)  2171--2175

\bibitem{Urbanowicz2012a}
Urbanowicz, R.J., Kiralis, J., Fisher, J.M., Moore, J.H.:
\newblock {Predicting the difficulty of pure, strict, epistatic models: metrics
  for simulated model selection}.
\newblock {BioData Mining} \textbf{5}(1) (2012)  1--13

\bibitem{Urbanowicz2012b}
Urbanowicz, R.J., Kiralis, J., Sinnott-Armstrong, N.A., Heberling, T., Fisher,
  J.M., Moore, J.H.:
\newblock {GAMETES: a fast, direct algorithm for generating pure, strict,
  epistatic models with random architectures}.
\newblock {BioData Mining} \textbf{5}(1) (2012)

\bibitem{Moore2013}
Moore, J.H., Hill, D.P., Sulovari, A., Kidd, L.C.:
\newblock {Genetic Analysis of Prostate Cancer Using Computational Evolution,
  Pareto-Optimization and Post-processing}.
\newblock In: Genetic Programming Theory and Practice X.
\newblock Springer New York, New York, NY (2013)  87--101

\bibitem{GiniImportance}
Breiman, L., Cutler, A.:
\newblock {Random forests - classification description} (November 2015)
  \url{http://www.stat.berkeley.edu/~breiman/RandomForests/cc_home.htm}.

\bibitem{Goldberg2002}
Goldberg, D.E.:
\newblock {The Design of Innovation: Lessons from and for Competent Genetic
  Algorithms}.
\newblock Kluwer Academic Publishers, Norwell, MA, USA (2002)

\bibitem{Konak2006}
Konak, A., Coit, D.W., Smith, A.E.:
\newblock {Multi-objective optimization using genetic algorithms: A tutorial}.
\newblock Reliability Engineering {\&} System Safety \textbf{91}(9) (2006)
  992--1007

\bibitem{Deb2002}
Deb, K., Pratap, A., Agarwal, S., Meyarivan, T.:
\newblock {A fast and elitist multiobjective genetic algorithm: NSGA-II}.
\newblock {IEEE Transactions on Evolutionary Computation} \textbf{6}(2) (2002)
  182--197

\bibitem{Greene2009}
Greene, C.S., Penrod, N.M., Kiralis, J., Moore, J.H.:
\newblock {Spatially Uniform ReliefF (SURF) for computationally-efficient
  filtering of gene-gene interactions}.
\newblock {BioData Mining} \textbf{2}(1) (2009)

\end{thebibliography}
\bibliographystyle{splncs}

\end{document}